%% file: main.tex
\documentclass[11pt]{article}

\usepackage{microtype}
\usepackage{graphicx}
\usepackage{multirow} 
\usepackage{booktabs} 
\usepackage{subfig}
\usepackage{longtable}
\usepackage{dblfloatfix}
\usepackage{amsmath}
\usepackage{wrapfig}

\usepackage{hyperref}

\usepackage{times}
\usepackage{latexsym}
\usepackage[final]{acl}

\usepackage{amsmath}
\usepackage{amssymb}
\usepackage{mathtools}
\usepackage{amsthm}
\usepackage{bbm}
\usepackage{longtable}
\usepackage{listings}

\usepackage{graphicx}

\definecolor{keywordcolor}{RGB}{86, 156, 214}  
\definecolor{stringcolor}{RGB}{214, 157, 133}  
\definecolor{commentcolor}{RGB}{87, 166, 74}   
\definecolor{numbercolor}{RGB}{255, 140, 0}    
\definecolor{backgroundcolor}{RGB}{30, 30, 30} 

\usepackage[capitalize,noabbrev]{cleveref}
\usepackage{xspace}

\newcommand{\our}{\textsc{UnDIAL}\xspace}

\theoremstyle{plain}

\theoremstyle{definition}

\theoremstyle{remark}

\usepackage[textsize=tiny]{todonotes}

\input{variables}

\title{
\our{}: Self-Distillation with Adjusted Logits for Robust Unlearning in Large Language Models
}

\author{\bf Yijiang River Dong$^{1}$, Hongzhou Lin$^{2}$, Mikhail Belkin$^{3}$, Ramon Huerta$^{2}$, Ivan Vulic $^{1}$ \\ \\
$^1$ University of Cambridge \ $^{2}$ Amazon \ $^{3}$ UCSD \\
}

\begin{document}
\maketitle

\begin{abstract}

Mitigating the retention of sensitive or private information in large language models is essential for enhancing privacy and safety. Existing unlearning methods, like Gradient Ascent and Negative Preference Optimization, directly tune models to remove unwanted information. However, these methods often become unstable because they fine-tune by maximizing cross-entropy loss, which is the opposite of traditional loss minimization in learning. This reversal creates instability, especially on larger datasets, as the model struggles to balance unlearning with maintaining language capacity, leading to over-unlearning. In this paper, we introduce \textsc{UnDIAL} (\textbf{Un}learning via Self-\textbf{Di}stillation on \textbf{A}djusted \textbf{L}ogits), a novel and robust unlearning method. Our approach leverages self-distillation to adjust logits and selectively reduce the influence of targeted tokens. This technique ensures smooth convergence and avoids catastrophic forgetting, even in challenging unlearning tasks with large datasets and sequential unlearning requests. Extensive experiments show that \textsc{UnDIAL} can achieve both robustness in unlearning and scalability while maintaining stable training dynamics and resilience to hyperparameter tuning.

\end{abstract}

\section{Introduction}
\input{sections/intro}

\section{Background and Related Work}
\label{s:rw}
\input{sections/related_work}

\section{Methodology}
\label{s:methodology}
\input{sections/methodology}

\section{Case Study One: Extraction Data}
\label{exp}
\input{sections/experiments}

\section{Case Study Two: MUSE Benchmark}
\label{s:muse}
\input{sections/experiment2}

\section{Conclusion}
\input{sections/conclusion}

\bibliography{main}


\newpage
\appendix
\section{Appendix}
\input{sections/appendix}

\end{document}

%% file: variables.tex
\newcommand{\rparagraph}[1]{\vspace{1.4mm}\noindent\textbf{#1}}

\newcommand{\sparagraph}[1]{\vspace{0.0mm}\noindent\textbf{#1}}

\newcommand{\rparagraphnodot}[1]{\vspace{1.4mm}\noindent\textbf{#1}}

%% file: sections/intro.tex

The increasing widespread use of large language models (LLMs) \cite{gpt4, MicrosoftCopilot2023, touvron2023llama, jiang2023mistral} in user-facing applications raises significant privacy concerns. Trained on vast, unmoderated web data, these models risk unintentionally exposing Personally Identifiable Information (PII), such as names and addresses \cite{heikkila2022does, white2023how}. Furthermore, LLMs are vulnerable to malicious exploitation, i.e. adversarial attacks \cite{carlini2021extracting, carlini2023quantifying, nasr2023scalable}, allowing confidential data to be extracted and heightening concerns about data security with AI \cite{levine2023generative}. 

In addition to these privacy risks, data protection regulations such as the EU's General Data Protection Regulation \cite{GDPR2016} and the California Consumer Privacy Act \cite{CCPA2018} enforce the ``right to be forgotten,'' enabling individuals to request the removal of their personal data from online platforms. This  creates an urgent need for techniques that allow LLMs to effectively ``unlearn'' and prevent the disclosure of specific information—a process known as \textit{LLM unlearning}.

Recent advances in LLM unlearning fall into two main categories. The first category involves using an \textit{auxiliary model} to explicitly memorize sensitive information, which is later removed from the original model using techniques such as contrastive decoding \cite{eldan2023whos, yu2021differentially, huang2024offset, ji2024reversing} or parameter merging \cite{ilharco_editing_2023, chen2023unlearn}. However, this approach introduces infrastructure overhead and poses a significant risk if the auxiliary model is exposed, as it contains exactly the data meant to be forgotten.

\begin{figure*}[t!]
\centering
\includegraphics[scale=0.47]{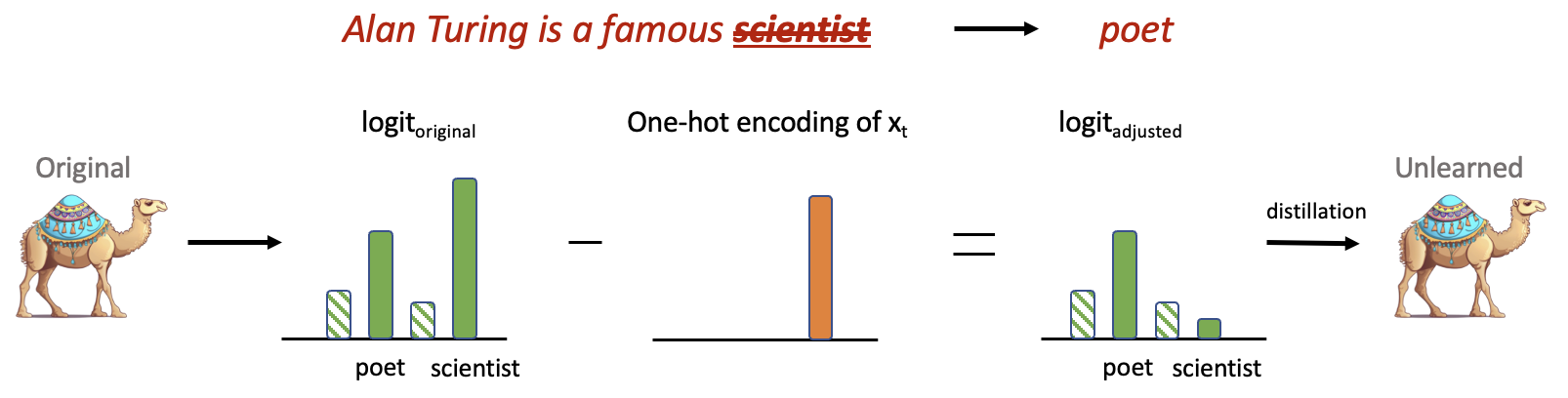}
\vspace{-1.5mm}
\caption{An illustration of the self-distillation process in the proposed \our{} method: The original logits generated by the model are adjusted by subtracting the one-hot distribution of the target token. The student model is then fine-tuned to approximate this modified logit distribution. Since the adjustments rely solely on the original model’s outputs, this is a self-distillation process to de-emphasize the token to be forgotten.
}
\vspace{-1.5mm}
\label{illustration}
\end{figure*}

Another line of research focuses on \textit{directly tuning} the base LLM model to unlearn sensitive information, using techniques such as Gradient Ascent (GA) \cite{jang2022knowledge} and Negative Preference Optimization (NPO) \cite{zhang2024negative}. These approaches are gaining more attention as they align more closely with the growing emphasis on AI safety \cite{gallegos2023survey, lucki2024adversarial}. 

Despite these advances, the recent unlearning benchmark MUSE \cite{shi2024muse} highlights a major drawback in current methods: applying unlearning to larger corpora leads to a decline in general language usefulness. This limits its usage in real-world settings, as an effective unlearning method must scale reliably with increasing data sizes, and accommodate continual updates—all while maintaining the model's overall language capabilities.




In this work, we introduce a novel direct-tuning method, \our{}, which enables Unlearning via Self-Distillation with Adjusted Logits. As shown in Figure~\ref{illustration}, we generate a target distribution by reducing the logit of the token to be unlearned. This target distribution is fixed during self-distillation, ensuring a stable optimization process. Unlike GA and NPO, which suffer from significant model capacity degradation as datasets scale and training extends, \our{} demonstrates strong robustness to data scaling, hyperparameter tuning, and sequential unlearning, offering the first robust unlearning method for direct tuning LLMs.

Our main contributions are as follows. \textbf{1)} We identify the robustness issues in current unlearning methods and propose a new, more robust method based on self-distillation. \textbf{2)} We demonstrate the effectiveness and robustness of \our{} across various hyperparameter settings,  forget set sizes and a number of unlearning requests. \textbf{3)} We also explore a variant of \our{} that focuses solely on specific set of tokens like named entities or nouns, which can further improve its overall performance.

%% file: sections/related_work.tex



\subsection{Memorization in Large Language Models} 

The LLMs' issue of memorizing and precisely reproducing data is getting increasingly recognized, especially when they get prompted in specific ways~\cite{carlini2021extracting, bender2021dangers, tirumala_memorization_2022, mccoy2023much}. This memorization behavior, while useful for encapsulating factual knowledge \cite{petroni2019language, Khandelwal2020Generalization}, also presents significant legal ramifications and challenges due to the unintended memorization of private or copyrighted material. Such instances increase the susceptibility of LLMs to extraction attacks or membership inference attacks \cite{carlini2021extracting, shokri2017membership,mireshghallah2022quantifying}. Recent studies have shown that, as these models grow in size, the dynamics of memorization fasten, leading to a linear increase in the fraction of data that can be extracted \cite{tirumala_memorization_2022, carlini2023quantifying}. To amortize such dynamics, techniques such as data deduplication \cite{lee2021deduplicating, kandpal2022deduplicating, nguyen2020variational} or private training are studied \cite{yu2021large, tramer2021differentially},  showing positive effect on reducing memorization. 



\subsection{Unlearning in Large Language Models}  \label{related_work}
Given the massive amounts of data involved in training LLMs, retraining these models each time to remove memorized data is   impractical. Thus machine unlearning focuses on how to effectively eliminate unintentional memorized content after the model is trained \cite{cao2015towards, ginart2019making, guo2020certified, bourtoule2021machine}. The unlearning algorithms can broadly fall into the following two categories: 



\rparagraph{Direct Tuning Methods.} \citet{jang2022knowledge} first formalize the problem of LLM unlearning and propose to use gradient ascent (GA) to achieve unlearning.  Instead of minimizing loss, GA maximizes the loss on tokens to be forgotten, forcing the model to forget specific knowledge. However, \citet{zhang2024negative} note that GA causes rapid collapse. They propose Negative Preference Optimization (NPO) which diverges slower than GA both in theory and practice. Alternative approaches tune the model to say ``I don't know" \cite{maini2024tofu} or predict random labels \cite{yao2024machine} on the knowledge that should be forgotten. 

\begin{figure*}[t!]
\centering
    \subfloat{
        \includegraphics[width=0.45\textwidth]{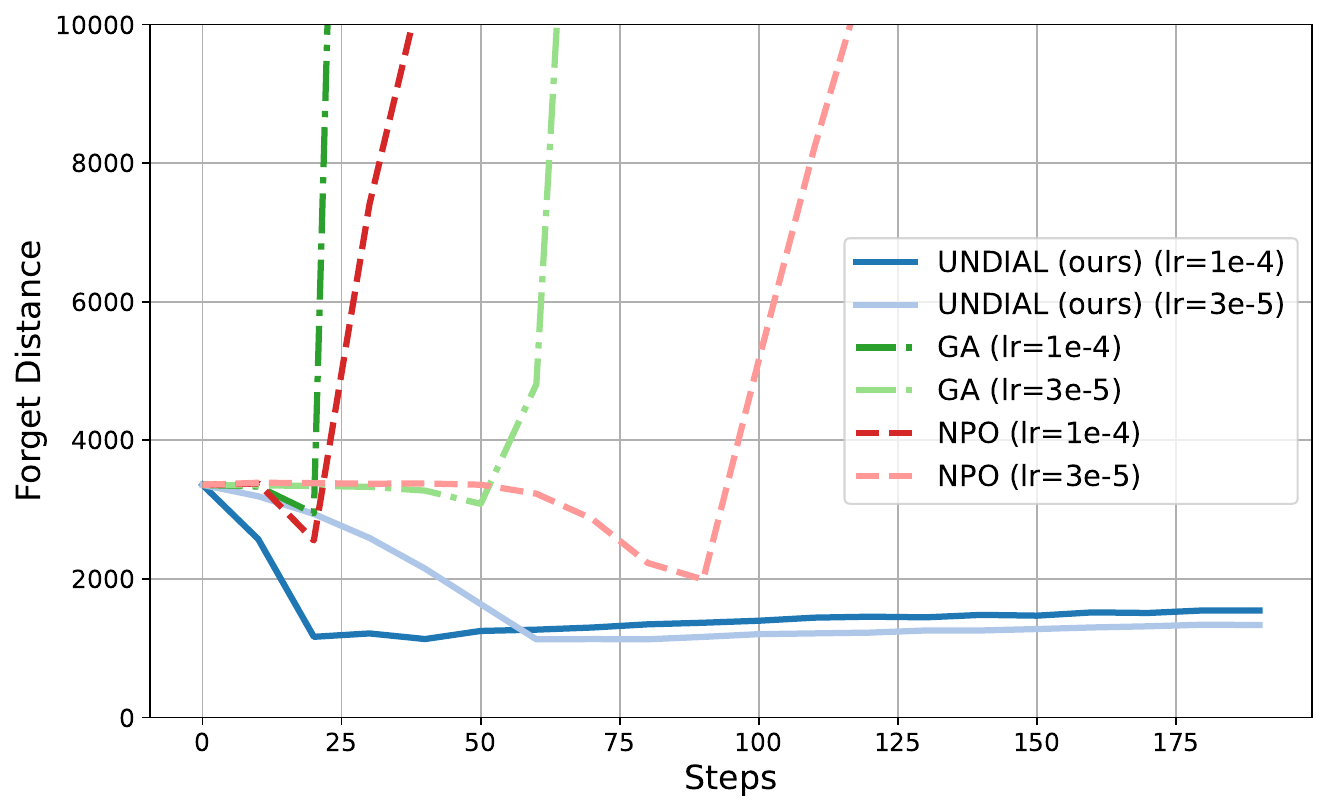}
    }
    \subfloat{
        \includegraphics[width=0.45
\textwidth]{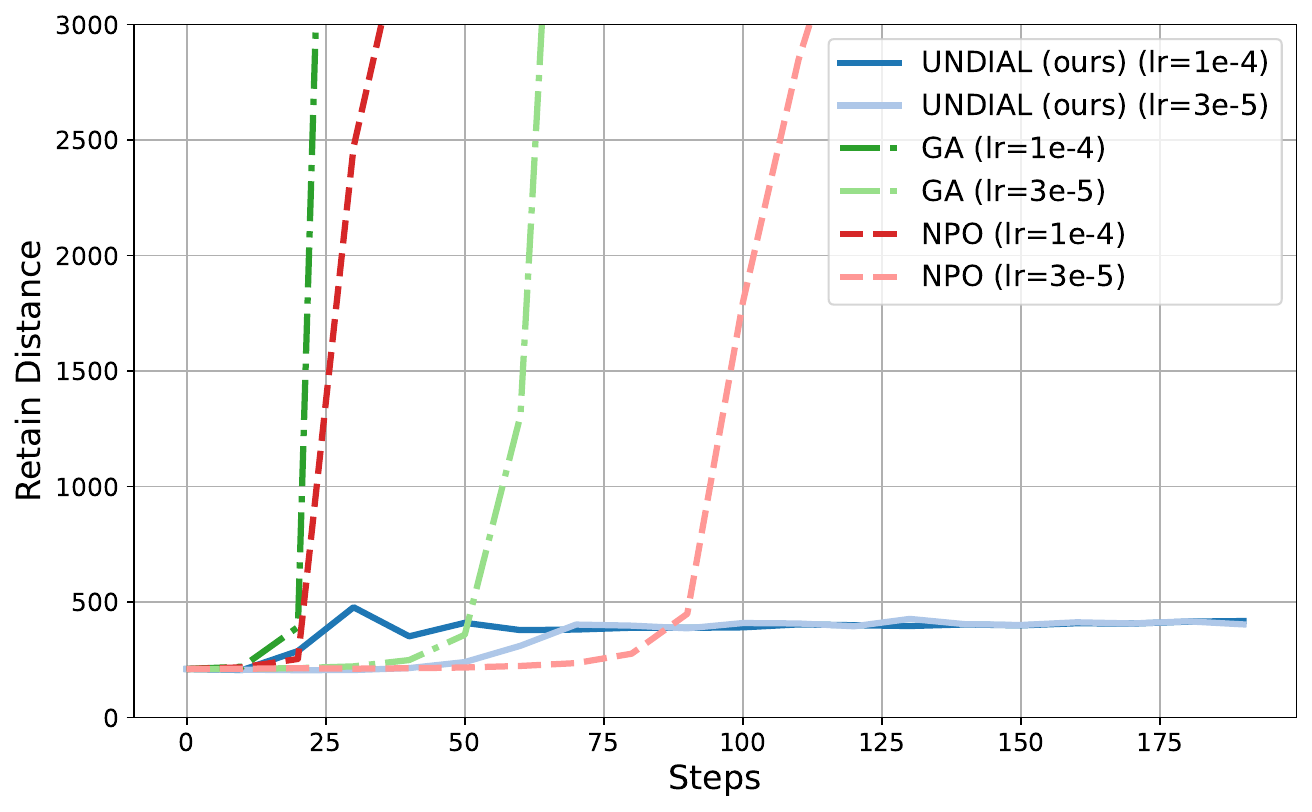}
    }
    \vspace{-1.5mm}
\caption{\textbf{Training dynamics of Direct Tuning methods on the MUSE benchmark \cite{shi2024muse}}. 
MUSE divides data into two sets: the Forget set, containing the information to be unlearned, and the Retain set, which measures the impact of unlearning on unrelated knowledge. Ideally, unlearning should be precise, affecting only the Forget set without disturbing the Retain set. MUSE provides fine-tuned models for both sets as optimal reference points. To capture the training dynamics, we compute the average KL divergence between the unlearned model and the MUSE reference models over the Forget and Retain sets. An effective unlearning model should closely match both references, with near-zero divergence indicating successful unlearning and model performance preservation.}
\label{forget_retain_distance}
\vspace{-1mm}
\end{figure*}

\rparagraphnodot{Leveraging Auxiliary Models} 
to remove the memorization in the base LLM, bypassing direct tuning, is in the focus of another line of research. \citet{eldan2023whos, ji2024reversing} first fine-tune a model to memorize the forget set and then leverage contrastive decoding (CD) \cite{li_contrastive_2023} to suppress the generation of unwanted memorization at decode time. Task Arithmetic (TA) approaches~\cite{ilharco_editing_2023} also fine-tune a model to memorize the forget set and leverage linear parameter merging~\cite{matena2022merging} to remove the memorization in model weights. \citet{majmudar2022differentially} apply linear interpolation with uniform distribution at the decoding time and show that this satisfies certain differential privacy (DP) criteria. \citet{chen2023unlearn} tune multiple unlearning layers to handle sequential unlearning requests and then fuse and plug them back into the base LLM.

We set aside post-processing methods such as directly \textit{prompting} LLMs to add a guardrail \cite{thaker2024guardrailbaselinesunlearningllms}; our focus is on removing knowledge directly from the base LLM via fine-tuning.





%% file: sections/methodology.tex




\sparagraph{Motivation with an Example.}  \label{motivation_subsection}
As highlighted by~\citet{zhang2024negative}, Direct Tuning methods face a major challenge of instability. Methods like GA and NPO, designed to directly unlearn from the original model, often lead to the so-called over-unlearning issue, referring to that unlearning algorithm is continuously applied to the model even after the corresponding knowledge is unlearned. This often leads to the model to collapse with the model capacity dropping to zero. While NPO partially mitigates this by adding a regularization term to slow the rate of divergence,  \textit{it fails to prevent long-term collapse in practice} \cite{fan2024simplicity}.

To demonstrate this critical issue, we apply GA and NPO methods on the MUSE dataset and illustrate the training dynamics in Figure \ref{forget_retain_distance}. As shown in the results, both GA and NPO exhibit model collapsing, although NPO diverges more slowly than GA. Through this empirical analysis we further corroborate the observation from recent related work~\citet{shi2024muse, fan2024simplicity},where both GA and NPO lead to over-unlearning and thus to a substantial decline in model usefulness and performance.

Moreover, NPO is also very brittle to different hyperparameter setups and thus difficult to tune. In the early stages of training, the distance on the forget set remains approximately constant (see Figure \ref{forget_retain_distance} left), showing that the model is not unlearning as expected. After this initial plateau stage, the model briefly begins to unlearn but quickly collapses. This instability reflects how sensitive NPO is to hyperparameter tuning and the need to stop training at exactly the right moment. Even slight overshooting can lead to severe performance degradation, i.e. over-unlearning. 

In contrast, foreshadowing our results, \our{} consistently shows robust performance throughout training, converging smoothly to a stable distribution. This stability allows for flexible stopping points without any risk of degradation. \our{} also achieves a substantially lower forget set distance in far fewer steps, making it \textbf{not just robust, but highly efficient}. As a Direct Tuning approach, we focus on comparing to GA and NPO in Figure~\ref{forget_retain_distance}. In Section~\ref{exp}, we will further show that \our{} outperforms even methods that rely on auxiliary models, proving its superiority across the board.

\subsection{\our{}: Method Description}

The main ingredient of our method is \textit{self-distillation}, where the model learns from its own predictions rather than external labels. Given an original model \( M_{original} \) and a sequence \( x_{1:T} \) that we aim to unlearn, the model generates a pre-softmax logit distribution at each token \( t \in [1, T] \):
\[
logit_{original} \sim M_{original}( \, \cdot \, | x_{<t}),
\]
representing a distribution over the vocabulary. To unlearn a specific token $x_t$, we reduce its logit value, forming an adjusted distribution:
\[
logit_{adjusted} \sim logit_{original} - \gamma e_{x_t},
\]
where \( e_{x_t} \) is a one-hot vector for token \( x_t \) and \( \gamma \) is a hyperparameter controlling the \textit{unlearning strength}. We then apply softmax to convert the adjusted logits into a probability distribution \( p_{adjusted} \), which de-emphasizes the tokens to be unlearned.

Since this adjustment is manually crafted, we perform self-distillation to learn the adjusted distribution by optimizing the model parameters $\theta$ so that $M_{\theta}$ can approximate the adjusted logits. This is done by minimizing the following loss function:
\[ L = \min_\theta \,\, \mathbb{E}_{x \sim D_{unlearn} }\left [\sum_{t=1}^T H( p_{adjusted} , p_{M_\theta} ) \right ] \]
where $H$ is the cross-entropy between the adjusted and model-generated distributions. As $p_{adjusted}$ is fixed, minimizing this loss corresponds to minimizing the KL-divergence between the two distributions, enabling the model to "forget" the specific tokens. 
In case of memorization, the token $x_t$ is typically the highest logit token among the entire vocabulary, i.e. $ x_t = \text{argmax}_{x \in \mathcal{V}} p_{original} (\, \cdot \, | x_{<t})$. To guide the model away from generating the memorized token, we subtract $\gamma$ from its logit, encouraging the model to generate the second-highest token instead. This reduces the probability of the memorized token; see again the example in Figure~\ref{illustration}.

\rparagraph{Why is \our{} Robust?} Unlike GA and NPO, which rely on maximizing loss, our method avoids the inherent instability via properly defining the target distribution. In GA and NPO, it is difficult to determine the optimal stopping point because the model lacks a clear convergence target. This often results in over-unlearning, instability, and eventual model degradation, especially when training is extended. The absence of a clear endpoint leads to a delicate balance between unlearning and retaining useful information, making these methods prone to catastrophic forgetting. In contrast, \our{} employs a well-defined target distribution that guides the model toward a stable outcome. This clear objective ensures smooth convergence, reducing the risk of over-unlearning and model degradation, and providing a robust, predictable optimization process. By focusing on a structured target, \our{} achieves both effective unlearning and the preservation of overall model performance.




\subsection{Variant: Focused \our{} (\textsc{FUnDIAL})}
\label{subsection: focus}
In the initial version of \our{}, self-distillation is applied uniformly across all tokens. However, not all tokens carry equal importance—some fulfill syntactic roles, while others, such as entity names and factual references, hold more critical information. For unlearning, it is more effective to apply stronger penalties to key tokens that encapsulate factual knowledge. Although identifying which tokens contain sensitive information can be subjective and challenging, we take a simple yet effective approach by treating nouns and entities as key tokens. This leads to a variant of our self-distillation method, where we adjust the distribution specifically for these key tokens. More formally, we introduce an entity indicator $\mathbbm{1}_{e}$ so that the loss for this variant only applies to specific targeted tokens:
\[ L_{f} = \min_\theta \,\, \mathbb{E}_{x }\left [\sum_{t=1}^T \mathbbm{1}_{e}(x_t) H( p_{adjusted} , p_{M_\theta} ) \right ].\]
In practice, we use the spaCy parser to extract entities and nouns, leaving further exploration of a more dynamical key token selection to future work.


%% file: sections/experiments.tex

\subsection{Dataset and Model} Following \citet{jang2022knowledge}, we use the dataset from the Training Data Extraction Challenge\footnote{\url{https://github.com/google-research/lm-extraction-benchmark}} to conduct unlearning. This dataset contains $15,000$ examples from the Pile dataset \cite{gao2020pile}, each consisting of $200$-token sequences. More importantly, the examples in this dataset have been proven to be memorized and are extractable from LLMs in the GPT-Neo family. This dataset is relatively smaller comparing to MUSE, allowing us to conduct extensive ablation studies. 

\subsection{Unlearning Metrics} 



\sparagraph{Memorization Accuracy (MA)} \cite{jang2022knowledge} measures the frequency of a given model~$M$ outputting the exact memorization tokens given the context, and it is computed as follows:
\begin{center}
$\rm{MA(x)} = \displaystyle \frac{\sum_{t=1}^{T-1} \mathbbm{1}[\rm{argmax}(p_{\theta}(\cdot  |x_{<t}) = x_t]}{T-1}$
\end{center}
%
\rparagraph{Extraction Likelihood (EL)} \cite{jang2022knowledge} generalizes the token level matching in the MA metric to $n$-gram overlap matching:
\begin{center}
    $\rm{EL_n(x)} = \displaystyle\frac{\sum_{t=1}^{T-n} \rm{Overlap_n}(M( \cdot |x_{<t}), x_{\geq t})}{T - n}$\\
    $\rm{Overlap_n}(a,b)$ = $\displaystyle\frac{|\rm{n\text{-}gram(a)} \cap \rm{n\text{-}gram(b)}|}{\rm{|n\text{-}gram(a)|}}$
\end{center}
Note that looping over all the context lengths from $1$ to $T-n$ is computationally expensive. We thus approximate MA and EL by only evaluating the overlap every $m$ tokens, i.e., on the context length as multiples of $m$. We set $m=40$. 



\subsection{`Model Usefulness' Metrics}
\label{ss:usefulness1}
In addition to unlearning metrics, we also evaluate general model usefulness via conventional Natural Language Understanding~(NLU) benchmarks and Generation (NLG) tasks.


\rparagraph{NLU Benchmarks and Metrics.} We measure the NLU capabilities by reporting the accuracy on six established NLU benchmarks: HellaSwag \cite{zellers2019hellaswag}, WinoGrande \cite{sakaguchi2021winogrande}, COPA \cite{gordon2012semeval}, ARC \cite{clark2018think}, PIQA \cite{bisk2020piqa}, and PubMedQA \cite{jin2019pubmedqa}. These evaluations are QA-style and are in line with the field's established best practices as used in previous studies~\cite{jang2022knowledge, eldan2023whos}.

\rparagraph{NLG Benchmarks and Metrics.} Here, we rely on the WikiText-103 datasets~\cite{merity2016pointer}. We select 5,000 samples from each dataset and use the first 32 tokens as a context prompt for the model to generate a continuation. The quality of these continuations is assessed using established open-generation metrics: MAUVE \cite{pillutla2021mauve} for semantic coherence, Repetition \cite{welleck2019neural} to check for redundancy, and Perplexity to evaluate overall fluency.

\begin{figure}[!t]
    \begin{minipage}{\linewidth}
        \includegraphics[width=0.93\linewidth]{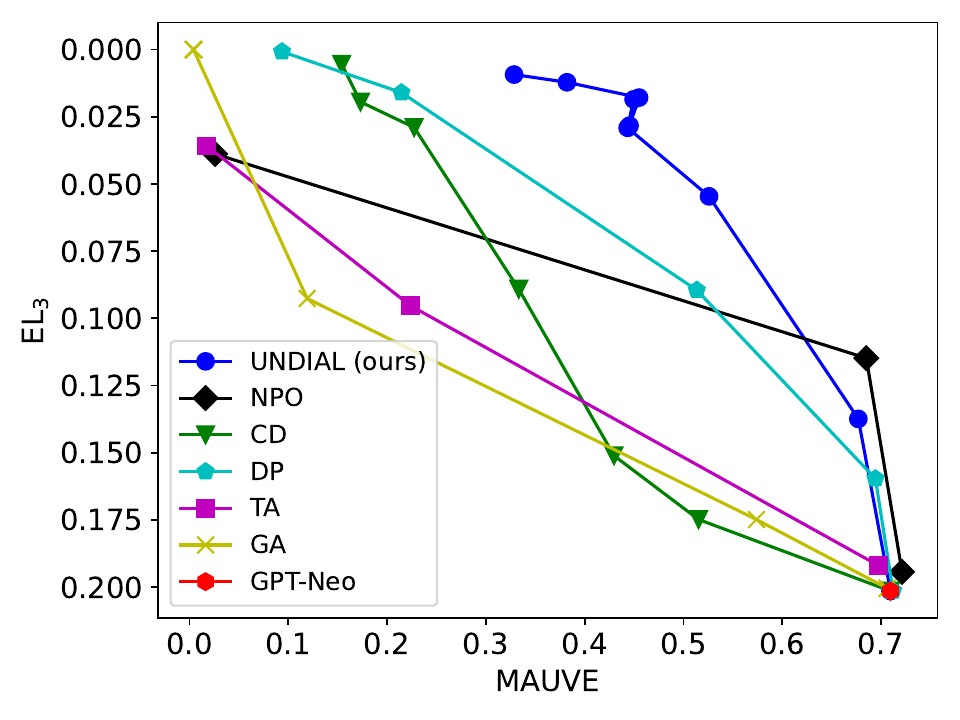}
    \end{minipage}\hspace{0.09\linewidth} 
    \begin{minipage}{\linewidth}
        \caption{\textbf{\our{} versus baselines when performing unlearning on the GPT-Neo 125M model.} The method with lower EL scores and higher MAUVE scores is considered better, i.e., towards the upper-right corner. For each of the methods, we vary the unlearning strength, naturally creating a curve of Pareto type showing the trade-off between memorization accuracy (EL) and language capacity (MAUVE).}
        \label{comparison}
    \end{minipage} 
    \vspace{-1cm}
\end{figure}

\subsection{Results and Discussion}
We now compare the results of our method~\our{} against all the representative baseline approaches such as Gradient Ascent (GA), Negative Preference Optimization (NPO), Differential Privacy (DP), Task Arithmetic (TA), and Contrastive Decoding (CD), outlined previously in \S~\ref{s:rw}. The experimental details can be found in Appendix~\ref{hyperparam_selection}, while a brief description of each baseline model is available in Appendix~\ref{appendix:baselines}.

\input{table/NLU_new}

\begin{figure*}[ht!]
\centering
    \subfloat{
        \includegraphics[width=0.3\textwidth]{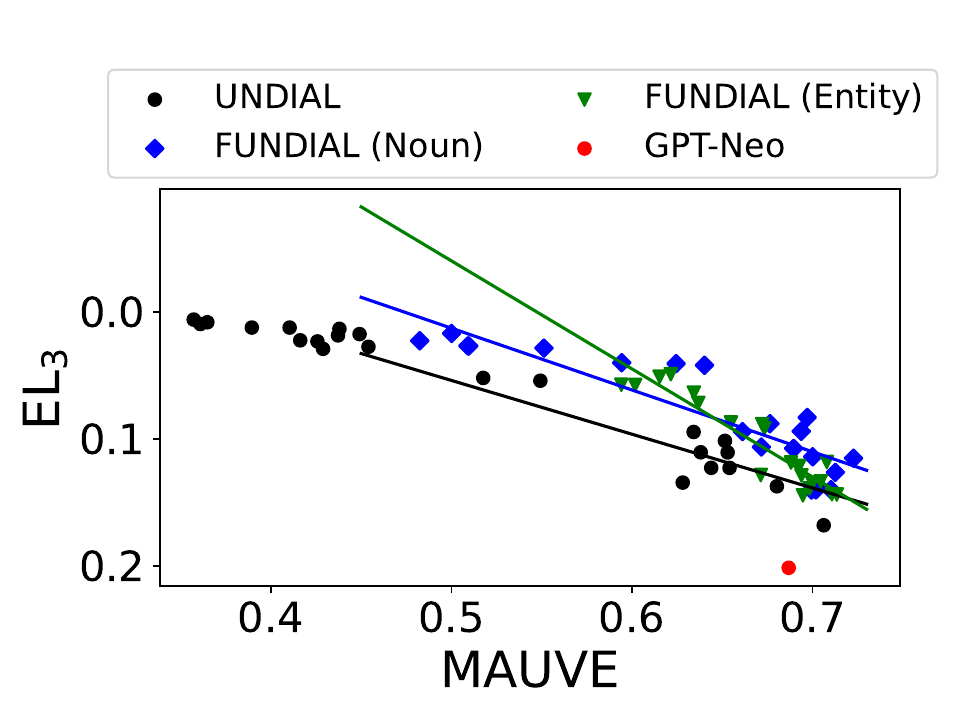}
    }
    \subfloat{
        \includegraphics[width=0.3\textwidth]{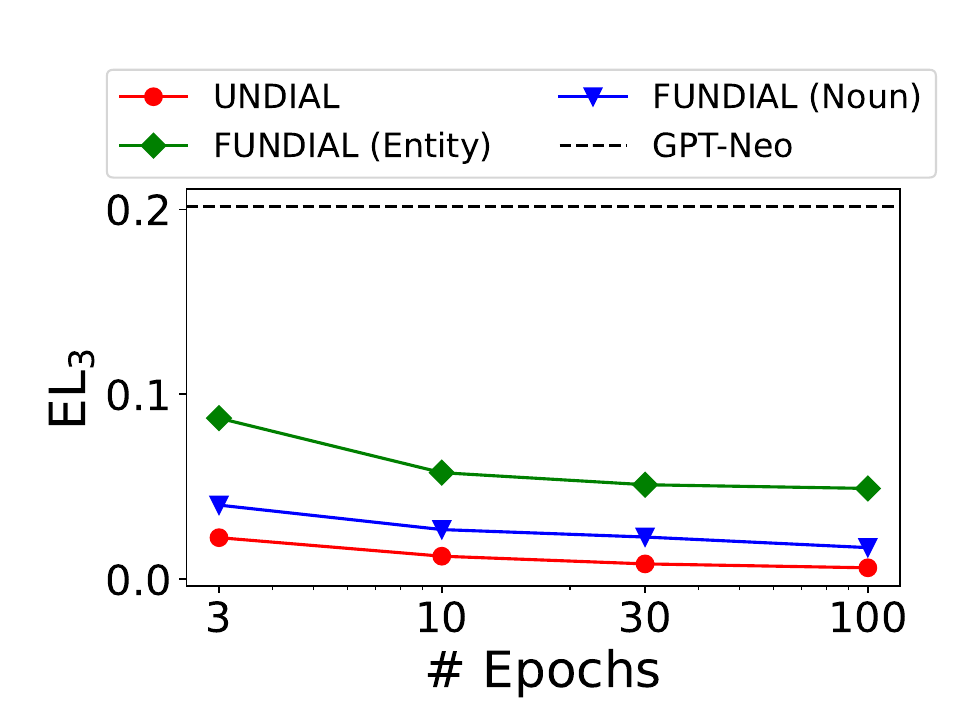}
        \label{focus_el}
    }
    \subfloat{
        \includegraphics[width=0.3\textwidth]{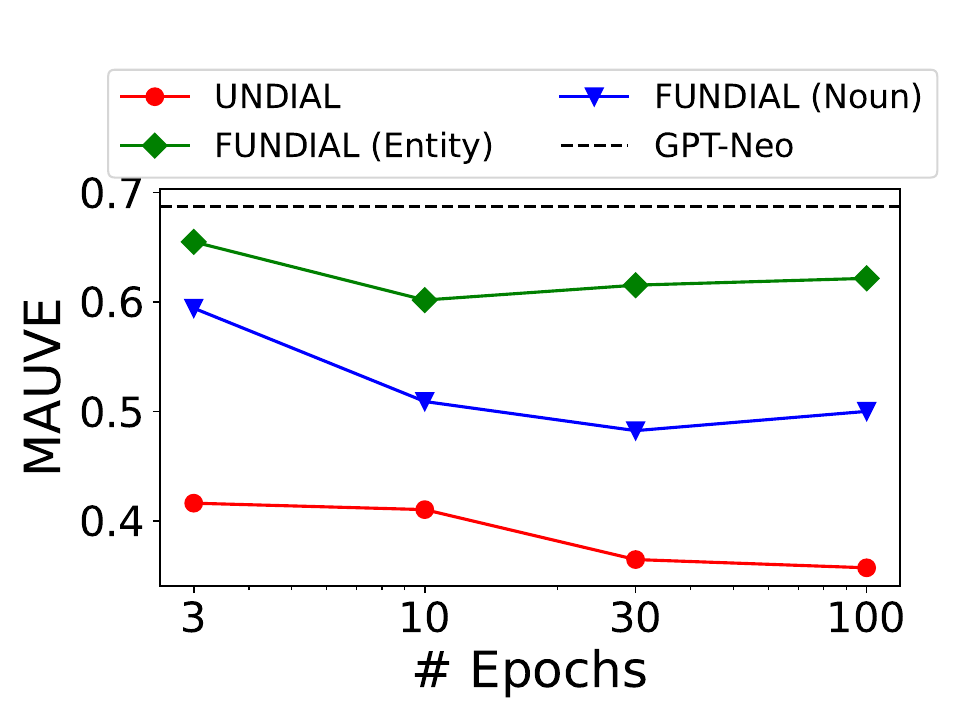}
        \label{focus_mauve}
    }
    \vspace{-1.5mm}
\caption{\textbf{Effectiveness of the Focused \our{} variant versus the basic variant}. The left figure shows the EL vs Mauve trade-off after introducing entity and noun indicators in our method, see \S\ref{subsection: focus}. By focusing on these specific tokens, we show that the performance can be further improved.   The two figures on the right show the stable training dynamics for a given unlearning strength $\gamma$=30 across different variants.}
\label{focus}
\vspace{-2mm}
\end{figure*}




\rparagraph{Unlearning versus Model Usefulness.} Figure~\ref{comparison} illustrates the trade-off between the memorization metric, Extraction Likelihood (EL), and the model usefulness metric, MAUVE. As we adjust the unlearning intensity for each method, a \textit{Pareto Frontier} naturally emerges, with the ideal point located in the upper-right quadrant. Our method excels here, achieving state-of-the-art language performance while maintaining high unlearning accuracy. The NPO method, while capable of comparable performance with careful tuning, quickly loses robustness as parameter settings change, demonstrating its sensitivity and lack of stability.


\rparagraph{{Full NLU \& NLG Evaluation.}} Table~\ref{NLU_new} presents results for QA-style NLU benchmarks and NLG tasks. Notably, NLG tasks are much more sensitive to unlearning, while NLU scores remain stable, within a $5\%$ margin from the GPT-Neo baseline. In contrast, NLG metrics show significant performance drops for several unlearning methods.

Focusing on rows with similar EL values around 0.1 (indicating a 50\% reduction in memorization), we observe that methods like GA, TA, DP, and CD degrade NLG performance significantly, as reflected in sharply lower MAUVE scores. Methods relying on auxiliary models (TA, DP, CD) perform worse on NLG tasks, showing a greater trade-off between memorization and usefulness. In contrast, NPO and our method \our{} maintain high MAUVE scores and experience less degradation in PPL and Rep\(_3\) metrics.




When reducing memorization further (EL $< 0.05$), NPO also sees a sharp decline in generation quality, with MAUVE scores falling below 0.03. We show examples in Appendix~\ref{examples}, where its outputs include repetitive or unnatural sentences. However, \our{} continues to generate high-quality outputs, even at these low memorization levels, highlighting its ability to balance unlearning and language generation quality.


This contrast in performance between \our{} and other methods underscores a critical insight in the field of LLM unlearning. While most existing methods tend to focus heavily on achieving unlearning at any cost, this often comes at the cost of neglecting and thus diminishing the model's language generation quality. Our method demonstrates that it is possible to achieve substantial unlearning (as evidenced by low EL scores) without sacrificing the quality of language output. Additionally, \our{} proves robust across different model sizes, as shown by the favorable scores of larger GPT-Neo variants (1.3B and 2.7B parameters) in Appendix~\ref{larger_gpt_neo}. This balance between effective unlearning and strong generation capabilities is essential for practical LLM applications.


\rparagraph{Focused~\our{}.} In the focused variant, we strategically fine-tune the model by focusing only on specific tokens, such as entity names or nouns, while not training on functional words. This targeted focus aims to improve the model's retention of language capabilities by avoiding the impact on the model predictions which concern functional words. The effectiveness of this method is shown in Figure~\ref{focus}. Our analysis reveals that, as the EL-MAUVE curve in Figure~\ref{focus} reveals, \textsc{FUnDIAL} outperforms the standard \our{}, which does not distinguish between different types of tokens. The position of \textsc{FUnDIAL} in the upper right corner of the curve suggests that a focused selection of targeted tokens leads to more effective unlearning and better preservation of language proficiency.

%% file: table/NLU_new.tex
\begin{table*}[t]
\centering
\small
\def\arraystretch{0.9}
\resizebox{\textwidth}{!}{
\begin{tabular}{lrrrrr rrrrrrl}
\toprule
& & & \multicolumn{3}{c}{\bf NLG Evaluation} & \multicolumn{7}{c}{\bf NLU Evaluation} \\
\cmidrule(lr){4-6}\cmidrule(lr){7-13}
\textbf{Method}                & Coeff & EL$_3$($\downarrow$) & MAUVE                        & PPL($\downarrow$)                & Rep$_3$($\downarrow$)         & PIQA                         & ARC                          & COPA                         & WinoG.                        & PubMed                       & HellaS.                       & Avg                          \\ \midrule
GPT-Neo (125M)              & -     & 0.202               & 0.718                        & 17.192                           & 0.035                        & 0.634                        & 0.383                        & 0.630                        & 0.515                        & 0.574                        & 0.282                        & 0.503                        \\ \midrule
                      & 0.05  & 0.117               & {\color[HTML]{FE0000} 0.305} & { 24.089}    & {\color[HTML]{FE0000} 0.174} & {\color[HTML]{009901} 0.613} & {\color[HTML]{009901} 0.365} & {\color[HTML]{009901} 0.680} & {\color[HTML]{009901} 0.521} & {\color[HTML]{009901} 0.575} & {\color[HTML]{009901} 0.279} & {\color[HTML]{009901} 0.504} \\
\multirow{-2}{*}{+TA}  & 0.10  & 0.035               & {\color[HTML]{FE0000} 0.017} & { 35.556}    & {\color[HTML]{FE0000} 0.515} & {\color[HTML]{000000} 0.560} & {\color[HTML]{000000} 0.291} & {\color[HTML]{000000} 0.560} & {\color[HTML]{000000} 0.514} & {\color[HTML]{000000} 0.535} & {\color[HTML]{000000} 0.264} & 0.454                        \\ \midrule
                       & 0.2   & 0.090               & 0.522                        & 76.428                           & 0.002                        & {\color[HTML]{009901} 0.611} & {\color[HTML]{009901} 0.345} & {\color[HTML]{009901} 0.546} & {\color[HTML]{009901} 0.516} & {\color[HTML]{009901} 0.571} & {\color[HTML]{009901} 0.277} & {\color[HTML]{009901} 0.478}                     \\
                      & 0.4   & 0.016               & {\color[HTML]{FE0000} 0.224} & {\color[HTML]{FE0000} 181.704}   & { 0.000} & 0.605                        & 0.320                        & 0.523                        & 0.521                        & 0.571                        & 0.266                        & 0.468                        \\
\multirow{-3}{*}{+DP}  & 0.6   & 0.001               & {\color[HTML]{FE0000} 0.082} & {\color[HTML]{FE0000} 308.882}   & { 0.000} & 0.601                        & 0.315                        & 0.539                        & 0.518                        & 0.571      & 0.261   & 0.468    \\ \midrule
                       & 0.25  & 0.089               & {\color[HTML]{FE0000} 0.333} & {52.202}    & { 0.056} & {\color[HTML]{009901} 0.611} & {\color[HTML]{009901} 0.346} & {\color[HTML]{009901} 0.644} & {\color[HTML]{009901} 0.516} & {\color[HTML]{009901} 0.576} & {\color[HTML]{009901} 0.278} & {\color[HTML]{009901} 0.495}\\
\multirow{-2}{*}{+CD}  & 0.5  & 0.017               & {\color[HTML]{FE0000} 0.172} & {\color[HTML]{FE0000} 158.187}    & { 0.042} & {\color[HTML]{009901} 0.592} & {\color[HTML]{009901} 0.319} & {\color[HTML]{009901} 0.630} & {\color[HTML]{009901} 0.504} & {\color[HTML]{009901} 0.562} & {\color[HTML]{009901} 0.273} & {\color[HTML]{009901} 0.480} \\ \midrule
                     & 1 & 0.174 & 0.573 & 15.133 & 0.053 & {\color[HTML]{009901} 0.622} & {\color[HTML]{009901} 0.367} & {\color[HTML]{009901} 0.630} & {\color[HTML]{009901} 0.514} & {\color[HTML]{009901} 0.575} & {\color[HTML]{009901} 0.283} & {\color[HTML]{009901} 0.499} \\

& 3 & 0.092 & {\color[HTML]{FE0000} 0.119} & 10.478 & {\color[HTML]{FE0000} 0.163} & {\color[HTML]{009901} 0.611} & {\color[HTML]{009901} 0.359} & {\color[HTML]{009901} 0.610} & {\color[HTML]{009901} 0.505} & {\color[HTML]{009901} 0.571} & {\color[HTML]{009901} 0.278} & {\color[HTML]{009901} 0.489} \\

\multirow{-3}{*}{+GA}                    & 5     & 0.000               & {\color[HTML]{FE0000} 0.004} & {3.381}     & {\color[HTML]{FE0000} 0.990} & 0.524                        & 0.257                        & 0.560                        & 0.498                        & 0.325                        & 0.258                        & 0.404                        \\ \midrule
                     & 1     & 0.114               & 0.685                        &   26.538                                & 0.077                        & {\color[HTML]{009901} 0.639}                        & {\color[HTML]{009901} 0.383}                        & {\color[HTML]{009901} 0.639}                        & {\color[HTML]{009901} 0.506}                        & {\color[HTML]{009901} 0.573}                        & {\color[HTML]{009901} 0.348}                        &  {\color[HTML]{009901} 0.515}                            \\
\multirow{-2}{*}{+NPO}                      & 2     & 0.038               & {\color[HTML]{FE0000} 0.026}                        &   17.683                               & {\color[HTML]{FE0000} 0.138}                       & 0.547                        & 0.284                        & 0.547                        & 0.507                        & 0.356                        & 0.283                        &     0.421                         \\  \midrule
                      & 3.0   & 0.111               & 0.674                        & 32.584                           & 0.010                        & {\color[HTML]{009901} 0.628} & {\color[HTML]{009901} 0.377} & {\color[HTML]{009901} 0.620} & {\color[HTML]{009901} 0.519} & {\color[HTML]{009901} 0.575} & {\color[HTML]{009901} 0.283} & {\color[HTML]{009901} 0.500} \\
                      & 10.0  & 0.019               & 0.450                        & 65.591                           & 0.005                        & {\color[HTML]{009901} 0.626} & {\color[HTML]{009901} 0.373} & {\color[HTML]{009901} 0.620} & {\color[HTML]{009901} 0.520} & {\color[HTML]{009901} 0.575} & {\color[HTML]{009901} 0.282} & {\color[HTML]{009901} 0.499} \\

\multirow{-3}{*}{+\our{} (ours)} & 30.0  & 0.013               & 0.437                        & 64.594                           & 0.005                        & {\color[HTML]{009901} 0.626} & {\color[HTML]{009901} 0.367} & {\color[HTML]{009901} 0.620} & {\color[HTML]{009901} 0.519} & {\color[HTML]{009901} 0.575} & {\color[HTML]{009901} 0.283} & {\color[HTML]{009901} 0.498} \\
\bottomrule
\end{tabular}
}%
\caption{\textbf{Performance of baseline methods and our \our{} method on NLU benchmarks and open-ended NLG tasks.} We highlight the NLU scores in green if the average accuracy decrease is less than 3\% and highlight the NLG scores in red if MAUVE drops more than half, or the repetition metric Rep$_3$ is above 0.1. Different methods control the unlearning strength via their own dedicated coefficients, which are detailed in Appendix \ref{appendix:baselines}. To interpret the table, we compare rows with similar EL values, such as those around 0.1, which indicates approximately a 50\% reduction in memorization.
}
\label{NLU_new}
\vspace{-20 pt}
\end{table*}

%% file: sections/experiment2.tex
We now evaluate our method on the MUSE dataset \cite{shi2024muse}, a recent, most comprehensive LLM unlearning benchmark. There, the data are coming from BBC News passages. The original work separates the data into two sets: \textit{Forget} and \textit{Retain}. In other words, we aim to unlearn a set of BBC News passages from the base model while retaining knowledge on other BBC News passages. 

For evaluation, question answering is conducted with respect to the News coming from the two sets. The questions related to the Forget set will test knowledge memorization, which we want to keep low. In contrast, the questions targeting the Retain set will test whether the unlearning procedure impacts unrelated topics, referred to as \textit{utility preservation}. Following the setup of~\citet{shi2024muse}, we use LLaMA-2 7B \cite{touvron2023llama} as the base model 
and LoRA~\cite{hu2021lora} with rank $r$ set to 8 to fit the fine-tuning onto one NVIDIA A100. 

\begin{figure}[t!]
\centering
\includegraphics[scale=0.26]{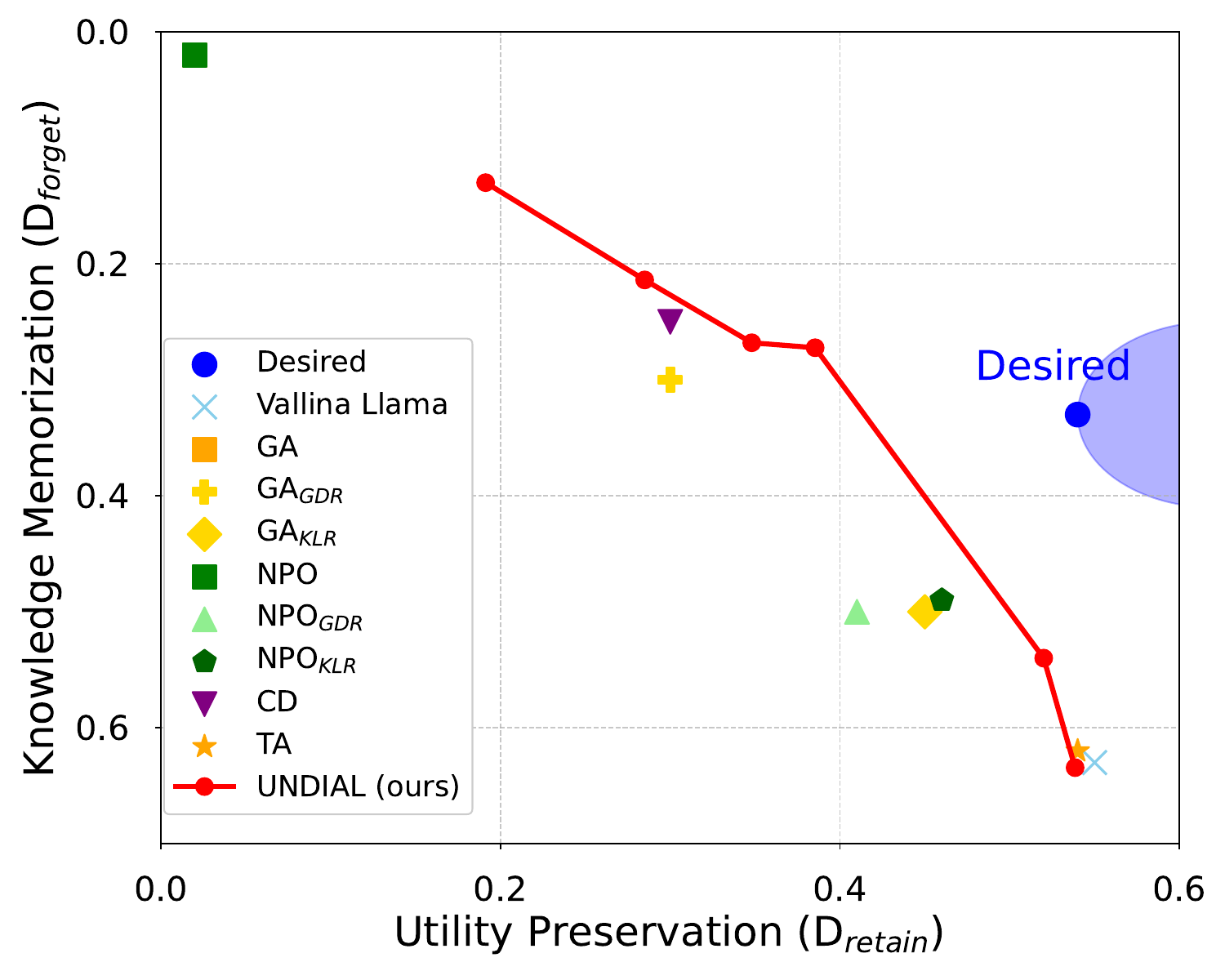}
\vspace{-1mm}
\caption{\textbf{Results on MUSE-News wih LLaMA-2 7B.} 
\textit{Knowledge Memorization} and \textit{Utility Preservation} refer to the accuracy on Q\&A with respect to the BBC News that aim to be forgotten and retained, respectively. The results of the baseline models are directly taken from~\citet{shi2024muse}. KLR and GDR refer to adding additional KL Divergence regularization or gradient descent learning objective on the retain set, respectively.}
\label{muse_tradeoff}
\vspace{-1.5mm}
\end{figure}

\rparagraph{\our{} Achieves a Better Pareto Frontier.} Figure~\ref{muse_tradeoff} shows the trade-off between model usefulness and unlearning achieved. By varying the unlearning strength, we observe that \our{} achieves a superior Pareto Frontier compared to the baseline methods, including both direct-tuning ones and the ones relying on auxiliary models (see~\S\ref{s:rw}).

Direct tuning methods like GA and NPO suffer from mode collapse, placing them near the origin. Some variants of those methods attempt to correct this by applying gradient ascent on the Forget set and gradient descent on the Retain set to balance the trade-off \cite{zhang2024negative,maini2024tofu}. While these adjustments help reduce mode collapse, they still underperform relative to \our{}.

Importantly, \our{} has the unique ability to achieve state-of-the-art performance without even relying on the Retain set at all. Unlike other methods that use the Retain set as additional information to help balance unlearning and general model usefulness, \our{} focuses solely on unlearning from the Forget set. This underscores the power of our approach: put simply, it achieves better results than methods that require extra data to maintain performance. 


\rparagraph{\our{} Shows Robust Training Dynamics.} Going back to the motivational example in Figure \ref{forget_retain_distance}, we reiterate that \our{} exhibits robust training dynamics while GA and NPO suffer from `over-unlearning' and catastrophic forgetting.



\begin{figure*}[t!]
\centering
    \subfloat{
        \includegraphics[width=0.39\textwidth]{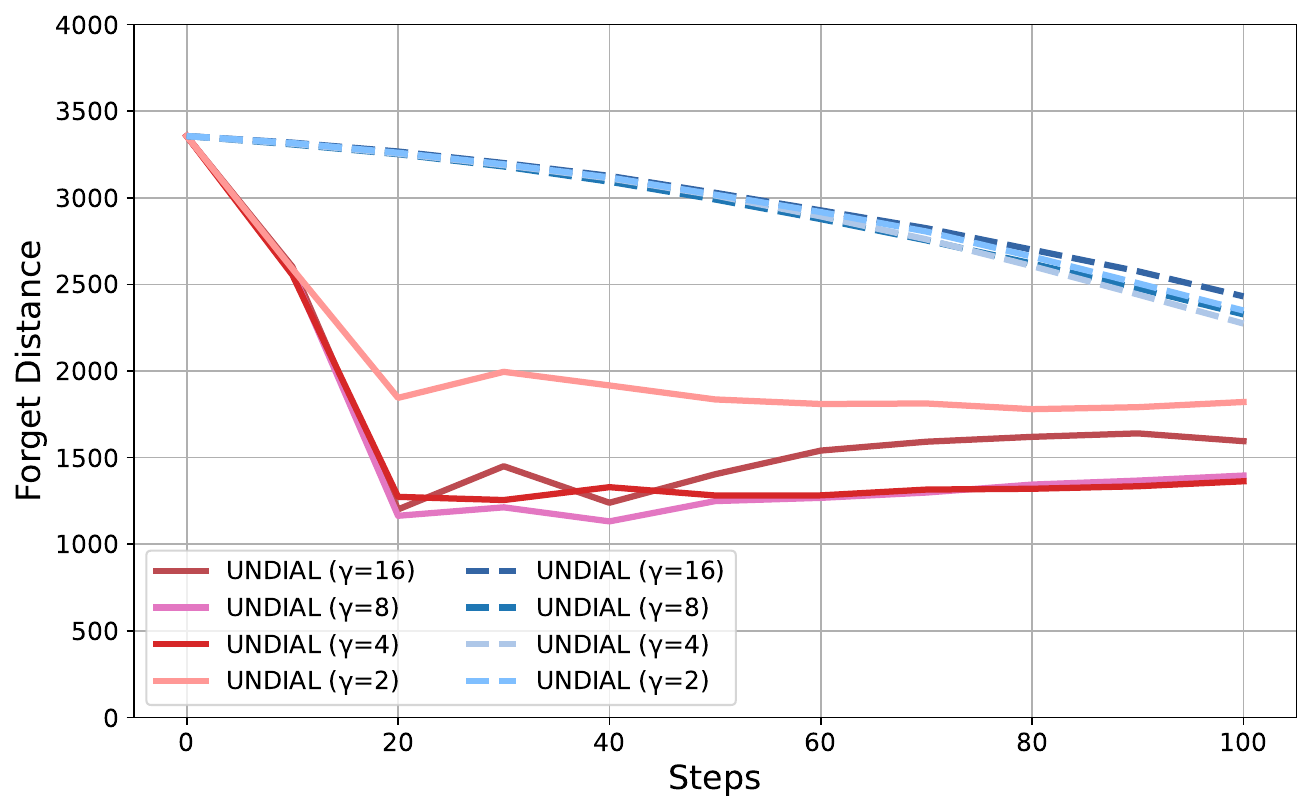}
    }
    \subfloat{
        \includegraphics[width=0.39\textwidth]{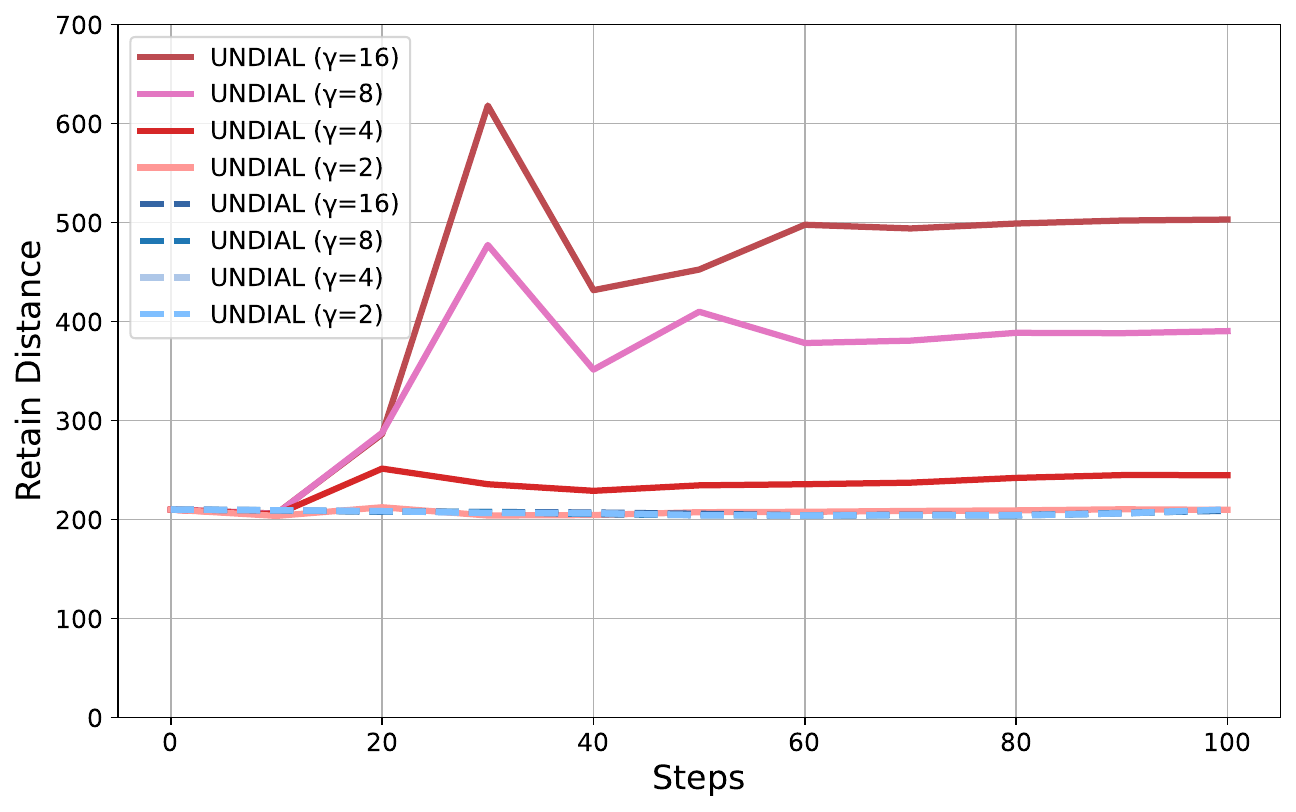}
    }
    \vspace{-1mm}
\caption{\textbf{Robust training dynamics of \our{} across different hyperparameter setups.} We show the training dynamics of our method with different learning rates (\textcolor{red}{red}: 1e-4, \textcolor{blue}{blue}: 1e-5) and unlearning strengths (\(\gamma = 2, 4, 8, 16\)). The forget and retain distances are measured by the KL divergence between the unlearned model and the optimal model from \citet{shi2024muse}. Unlike the unstable behavior of GA and NPO (see Figure~\ref{forget_retain_distance}), \our{} demonstrates stable training across all hyperparameter settings, confirming its robustness across different setups.
}
\label{forget_retain_distance_vary_strength}
\vspace{-1.5mm}
\end{figure*}

\rparagraph{\our{} is More Robust to Different Hyperparameter Setups.} In Figure~\ref{forget_retain_distance_vary_strength}, we illustrate the robustness of \our{} by varying the learning rate and unlearning strength~$\gamma$. We find that under different learning rates and unlearning strengths $\gamma$, the model still converges. However, it should be noted that, as expected, opting for more aggressive unlearning (i.e., increasing the unlearning strength) does hurt the model usefulness. For instance, in Figure~\ref{forget_retain_distance_vary_strength}(a), the forget distance of $\gamma$ set to 2, 4, 8 converges to a similar value while larger values for $\gamma$ lead to higher retain distances. However, while we observe some degradation and trade-off between unlearning and general model usefulness, unlike the other direct-tuning unlearning methods, \our{} does not suffer from the collapse issue, see Figure~\ref{forget_retain_distance} again.

\begin{figure*}[t!]
\centering
\includegraphics[scale=0.39]{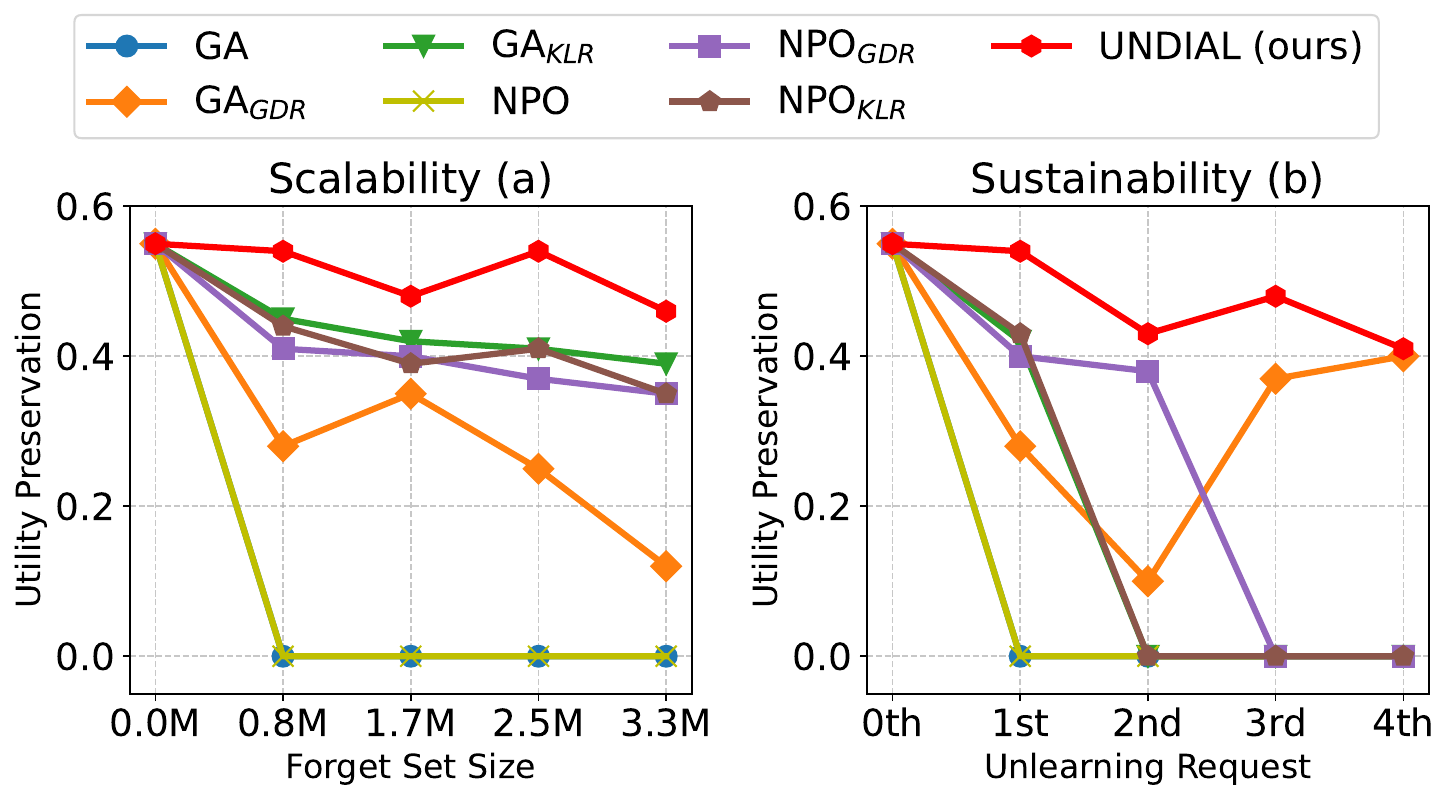}
\caption{\textbf{Robustness to (left) forget set size  and (right) sequential unlearning requests.} We conduct scaling and sequential unlearning tasks as done by~\citet{shi2024muse}. The baseline results on GA and NPO are taken from the original MUSE paper. In both tasks, \our{} is the most robust method and exhibits the best performance.}
\vspace{-1mm}
\label{muse_scale}
\vspace{-2mm}
\end{figure*}

\rparagraph{Unlearning with \our{} is More Scalable and Sustainable.} 
In real-world setups, the Forget set can become very large (the scalability feature) and the unlearning requests may come sequentially (sustainability). \citet{shi2024muse} test for these features and show that current unlearning methods are not robust to larger Forget set and sequential unlearning requests. However, we find that \our{} is much more scalable and sustainable than the baseline methods. Figure \ref{muse_scale}(left) shows that with larger Forget sets, model usefulness of \our{} is still reasonably well maintained, while GA and NPO's scores drop sharply. 

The experiments on sequential unlearning requests further demonstrate the robustness of \our{}. As shown in Figure \ref{muse_scale}(b), even as the number of unlearning requests increases, \our{} consistently maintains model utility above 0.4 with minimal degradation. In contrast, all baseline methods, including those with Retain set regularization, cause the model to collapse, with utility dropping close to zero as unlearning requests accumulate. 

This empirically validates that \our{} is able to avoid instability issues via properly defining the target distribution during unlearning. This ensures a more controlled and stable unlearning process. The robustness of our training dynamics is a key factor in maintaining model stability, particularly when scaling to larger datasets or handling sequential unlearning requests. 


%% file: sections/conclusion.tex
We introduced \our{}, a novel unlearning method based on self-distillation, which effectively balances reducing memorization with preserving language generation and understanding capabilities. Our approach represents a significant advancement in direct-tuning unlearning methods, offering improved robustness from multiple angles. Extensive experiments on the Extraction Data and MUSE benchmarks demonstrated state-of-the-art unlearning performance. Additionally, we show that \our{} is highly resilient across varying hyperparameters, different forget set sizes, and sequential unlearning requests. With its ability to prevent model collapse and scale efficiently, \our{} presents a promising next step for real-world applications requiring unlearning from LLMs.

\section*{Limitations}
We focus on a selected set of underlying language models (e.g., GPT-Neo and LLaMA-2 7B): this was motivated by their prior use on the same evaluation benchmarks coupled with the computational resources and budget available. Although these models already offer valuable insights into the unlearning performance of different approaches, we acknowledge that there is a possibility to extend the study to many other and larger LLMs in the future. We hope that this study will inspire other researchers and practitioners to port the main ideas behind \our{} to other model families and LLM architectures.


We also note that for the focused \textsc{FUnDIAL} variant of our approach, we take a reasonable yet very simplifying assumption on using only nouns and named entities as targeted tokens for the unlearning process. While empirically proven as effective, this approach may not always accurately identify the sensitive information and we envision more sophisticated approaches for the selection of focused tokens in future work. For instance, on potential improvement may be integrating an auto-detection mechanism for identifying privacy-sensitive data. This would enhance the method’s adaptability and ensure more comprehensive unlearning without relying solely on predefined classes of tokens.

\section*{Ethical Consideration}
Our paper introduces a novel method for addressing privacy concerns in LLMs. The approach aims to  enhance data privacy and security in LLM applications, aligning with broader societal needs for responsible AI. The societal consequences of improving privacy in LLMs are significant, potentially fostering greater trust and safety in LLMs used in various domains. Longer-term, we hope that models and initiatives focused on mitigating and removing concerns with how LLMs deal with private and sensitive data would also increase the (digital) society-wise trust in the (controlled) usefulness of such models.

%% file: sections/appendix.tex
\subsection{Hyperparameter Selection}
\label{hyperparam_selection}
We conducted our experiments using the GPT-Neo models~\cite{gpt-neo}, as they were used for extracting memorization data~\cite{carlini2021extracting}. We tested different model sizes: 125M, 1.3B, and 2.7B. Specifically, for the 125M model, we set the batch size to 64. For the larger 1.3B model, we used a mini-batch size of 16 and combined it with a gradient accumulation step of $4$ to make up the same $64$ batch size per gradient update. For all our experiments, we used the AdamW optimizer~\citep{adamw}.

\subsection{(A Brief Summary of) Baseline Models}
\label{appendix:baselines}

Multiple techniques have been recently proposed to address the unlearning challenge in LLMs, which we treat as the main baselines and briefly outline them in what follows. To describe the auto-regressive text generation process in models, we use the notation $x_t \sim p_\theta(\cdot|x_{<t})$, where $\theta$ represents model parameters, and $x_{<t}$ is the contextual information prior to position $t$.

\paragraph{Gradient Ascent (GA).} \citet{jang2022knowledge} introduce a technique that leverages memorized data identified from extraction attacks 
\citep{carlini2021extracting} to perform gradient ascent. This method decreases the probability of generating these memorized tokens by maximizing the log-likelihood loss on the memorized data, a reversal of the typical minimization approach: 
\begin{center}
$L_{UL} = - \sum_{t=1}^{T} log(p_\theta(x_t|x_{<t}))$
\end{center}
We vary training epochs in our experiments in Table \ref{appendix:baselines}.

\paragraph{Negative Preference Optimization (NPO)} \citet{zhang2024negative} treats the forget set as negative preference data and adapts the offline DPO objective to tune the model to assign low likelihood to the forget set without straying too far from the original model. Specifically, the NPO loss function becomes:

{\small
\[
\mathcal{L}_{\text{NPO}}(\theta) = -\frac{2}{\beta} \mathbb{E}_{x \sim \mathcal{D}_{\text{forget}}} \left[ \log \sigma \left( -\beta \log \frac{p_\theta(x_t | x_{<t})}{p_{\text{target}}(x_t | x_{<t})} \right) \right],
\]
}

where \( p_{\text{target}}(x_t | x_{<t}) \) refers to the target model probabilities, \( \sigma \) is the sigmoid function, and \( \beta \) controls the divergence from the target model \( f_{\text{target}} \). We set \( \beta = 0.1 \) in our experiments following \citet{zhang2024negative} and we vary training epochs in our experiments in Table \ref{appendix:baselines}.

\paragraph{Differential Privacy (DP).} Traditional DP methods \cite{bassily2014private, abadi2016deep} involve adding noise to gradients during the model training. However, the required noise level often scales with the number of parameters, leading to vacuous bounds for LLMs. 
 While more effective DP methods for fine-tuning have been suggested \cite{li2021large, yu2021differentially}, their performance discrepancies persist as the unlearning dataset increases \cite{anil2021large}. A more direct  baseline is to apply linear interpolation with uniform distribution at the decoding time, i.e.
\begin{center}
    $p(x_t|x_{<t})$ = softmax$((1-\lambda) z_t + \lambda u)$,
\end{center}
where $z_t$ represents the pre-softmax layer model output and $u$ is the uniform distribution over the vocabulary. 

\paragraph{Task Arithmetic (TA).} \citet{ilharco_editing_2023} apply 'task arithmetic' as a method for unlearning. This method fine-tunes a model on data to be forgotten and then subtracts these weights from the base model:
\begin{center}
$\theta_{TA} = \theta - \beta \cdot \theta_{memo}$, then $x_t \sim p_{\theta_{TA}}(\cdot|x_{<t})$
\end{center}
This coordinated subtraction requires the fine-tuned model to have the same architecture as the base model.

\paragraph{Contrastive Decoding (CD).} Similar to TA, contrastive decoding, as discussed by \citet{li_contrastive_2023} and further elaborated by \citet{eldan2023whos}, involves fine-tuning a model on data targeted for unlearning. The model's output probabilities are then adjusted either directly at the last layer or before it, incorporating an additional ReLU operation: 
\begin{center}
$p(x_t|x_{<t})$ = softmax$(z_t - \alpha \cdot z_t^{memo})$ 

\textbf{OR}

$p(x_t|x_{<t})$ = softmax$(z_t - \alpha \cdot \rm{ReLU}(z_t^{memo} - z_t)$)
\end{center}
where $z$ represents the pre-softmax layer model output  and $z^{memo}$ refers the fine-tuned model.  

\subsection{Scaling GPT-Neo on Unlearning Extraction Data}
In Table \ref{larger_gpt_neo}, we present the results of different sizes of GPT-Neo on the Extraction dataset. We validate that \our{} is robust across different model size from 1.3B to 2.7B and different fine-tuning methods, from full fine-tune to LoRA fine tune. 

\begin{table*}[t]
\def\arraystretch{0.88}
\centering
\resizebox{0.95\textwidth}{!}{
\begin{tabular}{l llllllllll}
\toprule & & & \multicolumn{4}{c}{\bf Unlearning Evaluation} & \multicolumn{3}{c}{\bf Language Capability Evaluation}
\\\cmidrule(lr){4-7}\cmidrule(lr){8-11}
\textbf{Method }    & \# Params & $\gamma$ & EL$_3$ & EL$_{10}$ & MA & Similarity   & MAUVE$\uparrow$ & PPL    & Rep$_3$ & NLU$_a\uparrow$ \\ \midrule
GPT-Neo    & 1.3B       & -        & 0.344     & 0.259          & 0.953 &  0.662 & 0.781 & 10.473 & 0.024      & 0.545    \\ \cmidrule(lr){4-11}
+\our{} (FT)   & 1.3B       & 3      & 0.111$_{\textcolor{red}{-0.233}}$     & 0.040          & 0.795 &  0.479   & 0.772$_{\textcolor{red}{-0.009}}$ & 12.685 & 0.024      & 0.543    \\
+\our{} (FT)   & 1.3B       & 10     & 0.070$_{\textcolor{red}{-0.274}}$     & 0.016          & 0.777 &  0.419   & 0.736$_{\textcolor{red}{-0.045}}$ & 15.288 & 0.021      & 0.546    \\ \cmidrule(lr){4-11}
+\our{} (LoRA) & 1.3B       & 3      & 0.091$_{\textcolor{red}{-0.253}}$     & 0.023          & 0.734 &   0.467    & 0.756$_{\textcolor{red}{-0.025}}$ & 12.516 & 0.030      & 0.543   \\
+\our{} (LoRA) & 1.3B       & 10     & 0.074$_{\textcolor{red}{-0.270}}$    & 0.015          & 0.712 &   0.424   &  0.723$_{\textcolor{red}{-0.048}}$ & 13.293 & 0.030      & 0.541    \\ \midrule \midrule
GPT-Neo    & 2.7B       & -                    & 0.389                & 0.309                & 0.966     &     0.695        & 0.800                & 9.442                & 0.024                & 0.582                \\ \cmidrule(lr){4-11}
+\our{} (LoRA) & 2.7B       & 3                 &      0.151$_{\textcolor{red}{-0.238}}$          &     0.067           &     0.803            &     0.525            &    0.795$_{\textcolor{red}{-0.005}}$           &    10.019         &      0.027           &  0.582  \\
+\our{} (LoRA) & 2.7B       & 10                 & 0.089$_{\textcolor{red}{-0.300}}$                & 0.022                & 0.768      &    0.467      & 0.774$_{\textcolor{red}{-0.026}}$                & 10.787               & 0.029                & 0.582                \\ \bottomrule

\end{tabular}
}%
\caption{\textbf{Results for different model sizes and with LoRA-based PEFT.} FT refers to full-model fine-tuning. We highlight the performance delta of EL$_3$ and MAUVE. NLU$_{a}$ is the average overall 6 NLU tasks. Lower is better, except with MAUVE and NLU$_a$.}
\label{larger_gpt_neo}
\end{table*}

\subsection{Implementation of \our{}}
We modified the typical huggingface Trainer with the following compute\_loss function.

\begin{figure*}[] 
\centering
\begin{lstlisting}
def compute_loss(self, model, inputs, return_outputs=False):
    input_ids = inputs['input_ids']
    attention_mask = inputs['attention_mask']
    student_logits = model(input_ids=input_ids, attention_mask=attention_mask).logits

    # Shift input_ids and logits for causal language modeling
    shift_labels = input_ids[..., 1:].contiguous()
    shift_student_logits = student_logits[..., :-1, :].contiguous()

    # Get teacher logits using the unlearned teacher model
    with torch.no_grad():
        teacher_logits = self.unlearn_teacher_model(
            input_ids=input_ids, attention_mask=attention_mask
        ).logits
    shift_teacher_logits = teacher_logits[..., :-1, :].contiguous()

    # Create mask for memorized tokens
    mask = torch.zeros_like(shift_student_logits)
    batch_indices = torch.arange(mask.shape[0]).view(-1, 1, 1)
    seq_indices = torch.arange(mask.shape[1]).view(1, -1, 1)
    mask[batch_indices, seq_indices, shift_labels.unsqueeze(-1)] = 1

    # Apply penalty to teacher logits and compute soft labels
    pre_softmax = shift_teacher_logits - mask * 10 # assuming a strength of 10
    soft_label = F.softmax(pre_softmax, dim=-1)

    # Compute cross-entropy loss between student logits and soft teacher labels
    loss_fct = CrossEntropyLoss(reduction='none')
    loss = loss_fct(shift_student_logits.view(-1, shift_student_logits.size(-1)),
        soft_label.view(-1, soft_label.size(-1)))
    return loss
\end{lstlisting}
\caption{Python Code for UNDIAL.}
\end{figure*}

\newpage
\onecolumn
\subsection{Generation Examples} 
In Table \ref{example_gen_forget} and Table \ref{example_gen_news} below, we show the generation example after unlearning.

\label{examples}

\subsubsection{Example Generation on CC News}
\input{table/examples_mauve}

\newpage
\subsubsection{Example Generation on Forget Set}
\input{table/examples_el}

%% file: table/examples_mauve.tex
\begin{table}[htbp!]
\small
\begin{tabular}{|c|c|c|p{9cm}|}
\hline
Method & Coeff & MAUVE & Text \\
\hline
GPT-Neo & - & 0.718 & use their weapons to help defeat or destroy enemies who attack their unit. If player will take control of their unit and use the weapons to take control of an enemy unit (or units) you can use this player in your campaign. It can be useful for people who don’t want to have to fight with the players in the game. This is how you can do it! https://docs.google.com/document/d/0Bz-3C4c-7R3/16E3C-zf8Pzv4M-1TJmS6J8/edit\#toc \\ 
\hline
UL & - & 0.004 & the the the the the the the the the the the the the the the the the the the the the the the the the the the the the the the the the the the the the the the the the the the the the the the the the the the the\\ \hline
TA & 0.05 & 0.305 &  use all tools of military force in it. The game provides an example of a military force attack against your team’s opponent’s player’s enemy. \newline
The game offers a gameplay of how to control Valkyria Chronicles game, a military Force attack against your team’s enemy’s player’s enemy and how to use all tool that use Valkyria Chronicles IV.\newline
With the advent of Valkyria Chronicles games, Valkyria Chronicles III : the game begins with the game about how to use a Valkyria Chronicles III. The game offers a gameplay of how to use a Valkyria Chronicles III. The game features you can choose your team’s enemy
 \\ \hline
\multirow{3}{*}{DP} & 0.6 & 0.005 &  explore the world using a top-down viewpoint on an alien world filled with action. There’s also side-quests, new classes, units, and weapons to collect. This release allows fans around Xbox 360 / Xbox one users access to one playable mission or 12 side quests across both factions of this campaign (Vidya vs Zefran, Zephyr VS Zair or Vidya VS Zerro). \newline
In addition, the new update allows for better graphics to provide a high fidelity and detail experience. \newline
You, your companions, and your friends who decide to work with your unit have more choices, options and story events with all gameplay features on all playable campaign missions and 12 sidequest across both nations of this campaign (Vidya vs Zephyr, Zair Vs Zerro) \\ \cline{2-4} 
& 0.4 & 0.082 & interact to create and/or protect special elements / combat system of attack. On top we also have an advanced combat engine! (In earlier series) This title was created for your needs – more games! \*The games feature one main screen – `View Games/Wards (Game Tools/Video Games): Full size view for one (full). In both models this functionality can allow two groups play multiple areas; one group `play’ on ``front” part with three dots“down and one for [focalsis on ``ficos da carregavagno [game assets]]; the left end [full for group playing’], whilst another for [back to player] in ``the middle screen", playing group with dots.[c1e]The basic screen aspect of all V.E.- and NFSO III was designed to provide your characters or groups from one game can display on another
\\ \cline{2-4} 
& 0.2 & 0.224 &  train other players throughout timeframes which change time intervals, and ultimately they manage those different days and allow players at many moments into hours.
 \\ \hline
\end{tabular}
\end{table}

\begin{table}[htbp!]
\small
\begin{tabular}{|c|c|c|p{9cm}|} 
\hline
Method & Coeff & MAUVE & Text \\
\hline
CD & 0.25 & 0.172 &  build a fleet. \newline
There is no more exciting way to play Valkyria Chronicles 3, and unlike previous Valkyria games, the gameplay is more of a tactical game, as players are able to build the fleet and get back to the beginning of the game with no issues.
 \\ \cline{2-4} 
   & 0.5 & 0.333 &  a command centre and a command station, and play in a team based game, which includes the following roles : \newline
Masters of the Universe (MOV) - The Master of the Universe (MOV) is an ancient weapon of the Galactic Empire, which can be used in a variety of different ways. For example, it can be used in a variety of different ways, su ch as using the ability to use a variety of weapons. \newline
Masters of the Universe (MOV) - The Master of the Universe (MOV) can be used in a variety of different ways. 
 \\ \hline
\multirow{3}{*}{DI (Ours)} & 3 & 0.674 &  use their equipment to control their operations and take control from others to play their part in their respective roles. The games of Valkyria Chronicles 3 can take in a variety of forms, but in every role is its very particular role and its main goals can take in a variety of roles - from military, to industrial - but its primary focus is its role as player and the only player roles are the most specific of both combat and tactical. \\ \hline
 \\ \cline{2-4} 
   & 10 & 0.450 &  use their equipment and equipment to perform operations that require an individual to play their part. This style gameplay mode is very different when compared against previous two of these in series. In each season in V-4 we have a single player role in play with the only player who can take control - a game mechanic (vulnerable in its abilities to be a combatant in an emergency), with a single command of three or four combat aircraft with which you can defeat a unit. However this only applies in a two game-by two game mode with the same amount or combination to each player. You will need a team of three or four players for this style gameplay to be effective at this point in any game. As for a four player team when a two is needed and a five player game for this side you can do the same as you have in previous V – but this requires several players not on the defensive.\\ \hline
   
Prompt & \multicolumn{3}{p{12cm}|}{As with previous Valkyria Chronicles games, Valkyria Chronicles III is a tactical role playing game where players take control of a military unit and} \\ \hline
Gold Continuation & \multicolumn{3}{p{12cm}|}{take part in missions against enemy forces. Stories are told through comic book-like panels with animated character portraits, with characters speaking partially through voiced speech bubbles and partially through unvoiced text. The player progresses through a series of linear missions, gradually unlocking maps that can be freely scanned through and replayed as they are unlocked. The route to each story location on the map varies depending on an individual player's approach: when one option is selected, the other is sealed off to the player. Outside missions, the player characters rest in a camp, where units can be customized and character growth occurs. Alongside the main story missions are character-specific sub-missions relating to different squad members.} \\ \hline

\end{tabular}
\end{table}
\label{example_gen_news}

%% file: table/examples_el.tex
\begin{table}[htbp]
\caption{Some examples generated by GPT-Neo 125M after doing unlearning. We highlight the text that are direct memorization of the training data.}
\vskip 0.15in
\small
\begin{tabular}{|c|c|c|p{9.5cm}|} 
\hline
Method & Coeff & EL$_3$ & Text \\
\hline
GPT-Neo & - & 0.202 &  
\textcolor{red}{Odoo Server saas~}10."\newline
\textcolor{red}{"Report-Msgid-Bugs-To: "}\newline
\textcolor{red}{"POT-Creation-Date: 201}8-09-21 09:59+0000"\newline
\textcolor{red}{"PO-Revision-Date: 201}8-09-21 09:59+0000"\newline
\textcolor{red}{"Last-Translator: Martin Trigaux $<$mat@odoo.com$>$"}\newline
\textcolor{red}{"Language-Team:} Hebrew \textcolor{red}{(https://www.transifex.com/odoo/teams/\newline41243/he/)"}\newline
\textcolor{red}{"MIME-Version: 1.0"}\newline
\textcolor{red}{"Content-Type: text/plain; charset=UTF-8"}\newline
"Content-Transfer-Encoding: 8bit"\newline
"Language: he"\newline
"Plural-Forms: nplurals=4; plural=(n==1? \\
\hline
UL & - & 0.000 & the the the the the the the the the the the the the the the the the the the the the the the the the the the the the the the the the the the the the the the the the the the the the the the the the the the the the the the the the the the the the the the the the\\ \hline
TA & 0.05 & 0.117 & -16 02:41-08-15 17:01-07-15 - - "\newline
"Language: English"\newline
\textcolor{red}{"Content-Type: text/plain; charset=}ISO-8859-1"\newline
"MIME-Version: 1; charset=ISO-8859-1"\newline
"Content-Transfer-Encoding: 8bit"\newline
"Expires-By: "\newline
"Content-oji-Date: 2016-09-05 08:00-06-03 10:30-04 10:30"\newline 
"Content-oji-Progetty: " \\ \hline
\multirow{3}{*}{DP} & 0.6 & 0.090 &   Package"\newline
"Report-Msgid-Bugs-To"\newline
\textcolor{red}{"POT-Creation-Date:} 2020-0500220000"\newline
"PO-Revision-Date: 2019-08-12 15:56+0000"\newline
\textcolor{red}{"Last-Translator: Martin Trigaux $<$mat@odoo.com$>$"}\newline
``Language-Team: Portuguese (PTB)"\newline
``Language: pa-PUBA"\newline
``MIME-Version: 0 to 64)"\newline
\textcolor{red}{``Content-Type: text/plain"}\newline
``Content-Transfer-Encoding: 8bit" \\ \cline{2-4} 
   & 0.4 & 0.016 &  projectlister "\newline
``PO-Description-Tr-LanguageLists:(u 18 z/b) "\newline
``Project-Target- matrix"\newline
""\newline
\#: projectdists/contr-5.0+x\_10\_12.5 xmlcontent.cfm\newline
``$<$?xml$>$$\rm{{projectxml.content2version-1.0|2}}"\newline //${version}.txt"\newline$>$$<$tr$>$$<$row$>$$<$col width="18dp"$>$$<$col $>$/1$>$$<$row$>$$<$li $>$$<$p$>$Name (if this are correct, just add $<$li$>$before 'name(width)+100' (the text) is always black when all data and controls are black ($<$br$>$name in case, you added other text below if applicable) or in all circumstances ('other text is a red color as with previous font types; and you do care because this color
 \\ \cline{2-4} 
& 0.2 & 0.001 & Odev O'De Landing (2+5) Console UI (\%1=Overtie Voorwaar volderen) - DER cplication vf/c2n vcf (1 x\%24m - \%4a ) \%17/12a \"Nummer\" vidierstrignt \%4a \%12m01
 \\ \hline
\end{tabular}
\end{table}

\begin{table}[ht]
\small
\begin{tabular}{|c|c|c|p{8.5cm}|} 
\hline
Method & Coeff & EL$\_3$ & Text \\
\hline
CD & 0.25 & 0.089 & .dists"\newline
"Report-Msgid-Bugs-To: "\newline
\textcolor{red}{"POT-Creation-Date: 2016}-02-04 14:30+0200"\newline
\textcolor{red}{"PO-Revision-Date: 2017}-01-21 12:43+0000"\newline
\textcolor{red}{"Last-Translator: Martin Trigaux <mat@odoo.com>"}\newline
\textcolor{red}{"Language-Team:} Danish"\newline
"Language: dde"\newline
\textcolor{red}{"MIME-Version: 1.0"}\newline
\textcolor{red}{"Content-Type: text/plain; charset=UTF-8"}\newline
"Content-Transfer-Encoding: 8bit"\newline
"X-Generator: Launchpad zendump"\newline
 \\ \cline{2-4} 
   & 0.5 & 0.017 &  Odoo Serverifle"\newline
"Report-Msgid-Bugs-To: "\newline
\textcolor{red}{"POT-Creation-Date:} 2018-09-08 02:08+0000"\newline
"PO-Revision: 1. znal@postazion.de"\newline
\textcolor{red}{"Last-Translator: Martin Trigaux $<$mat@odoo.com$>$, 2017}"\newline
"Language-Team: Greek"\newline
"Language: Greek"\newline
"MIME-Version: 1.0"\newline
\textcolor{red}{"Content-Type: text/plain; charset=UTF-8"}\newline
"Content-Transfer-Encoding: 8bit"\newline
"Plural-Forms: nplurals=3; plural=(n==1? 0 : n== counts*number)? 1:"
 \\ \hline
DI (Ours) & 3 & 0.111 &  P4-LitP4.1.8"\newline
"Report-Msgid-Branch: "\newline
"Phing-Version: 1.0"\newline
"Language: en"\newline
\textcolor{red}{"MIME-Version: 1"}\newline
"Content-Language: en"\newline
\textcolor{red}{"MIME-Type: text/plain}; e"\newline
"Plural-Language: en"\newline
"Generated-by: The GORO"\newline
"Date-WkD-Version: 2018-02-23"\newline
"For-Wkt-Relevant-PkGx-Version: 2"\newline
\\ \hline

DI (Ours)   & 10 & 0.019 &  Odoo Project"\newline
"PO-Revision-Date: 20190517165900
"Generated"\newline

\#. Translation: For translations which are based on local content,
\# this message is translated by the team responsible for the localization of Odoo (the translation team is called
\# \"Global-Translator\" in Odoo source repositories). Note: All the other content of the message should be replaced by the corresponding local translation team which is called $<$translatorid$>$. However, this is not always possible, and this message is just not included in Odoo source documentation. Therefore, we decided to translate this message. \newline
msgid "" \newline
msgstr "" \newline
\#. Translators: Odoo language: Odoo project $<$projectid$>$
msgid "Translators: $<$project-number@odoo.com$>$ $<$translator-number$>$. $<$group-number$>$. Odoo Language Team $<$teamname$>$."
\\ \hline
Prompt & \multicolumn{3}{p{10cm}|}{\#  \newline 
\# Translators: \newline 
\# Martin Trigaux $<$mat@odoo.com$>$, 2017 \newline 
msgid "" \newline 
msgstr "" \newline 
``Project-Id-Version:} \\ \hline
Gold Continuation & \multicolumn{3}{p{10cm}|}{Odoo Server 10.saas~18"\newline
``Report-Msgid-Bugs-To:" \newline 
``POT-Creation-Date: 2017-09-20 09:53+0000"\newline
``PO-Revision-Date: 2017-09-20 09:53+0000"\newline
``Last-Translator: Martin Trigaux $<$mat@odoo.com$>$, 2017"\newline
``Language-Team: Kabyle (https://www.transifex.com/odoo/teams/41243/kab/)"\newline
``MIME-Version: 1.0"\newline
``Content-Type: text/plain; charset=UTF-8} \\ \hline

\end{tabular}
\end{table}
\label{example_gen_forget}